\documentclass[12pt]{article}
\usepackage{times}

\usepackage{amsmath}
\usepackage{amssymb}
\usepackage{float}
\usepackage{graphicx}

\usepackage[square,sort,comma,numbers]{natbib}
\usepackage{color} 

\usepackage{url}
\usepackage{pdfpages}
\usepackage{setspace} 

\usepackage{hyperref}

\hypersetup{
	bookmarks=true,         
	unicode=false,          
	pdftoolbar=true,        
	pdfmenubar=true,        
	pdffitwindow=false,     
	pdfstartview={FitH},    
	pdftitle={My title},    
	pdfauthor={Author},     
	pdfsubject={Subject},   
	pdfcreator={Creator},   
	pdfproducer={Producer}, 
	pdfkeywords={keywords}, 
	pdfnewwindow=true,      
	colorlinks=true,       
	linkcolor=blue,          
	citecolor=blue,        
	filecolor=magenta,      
	urlcolor=cyan           
}

\usepackage[labelfont=bf,labelsep=period,justification=raggedright]{caption}

\makeatletter
\renewcommand{\@biblabel}[1]{\quad#1.}
\makeatother

\date{}

\title{\bf{Artificial Intelligence Development Races in Heterogeneous Settings}}

\author{Theodor Cimpeanu$^{1}$, Francisco C. Santos$^{3}$, Lu\'is Moniz Pereira$^{2}$, \\ Tom Lenaerts$^{4,5,6,7}$, and The Anh Han$^{1,\star}$}

\begin{document}
	\maketitle
	{\footnotesize
		\noindent
		$^{1}$ School of Computing, Engineering and Digital Technologies,  Teesside University, Middlesbrough, UK TS1 3BA\\
		$^{2}$  NOVA Laboratory for Computer Science and Informatics (NOVA-LINCS), Faculdade de Ci\^encias e Tecnologia, 
		Universidade Nova de Lisboa, 
		2829-516 Caparica, Portugal \\
		$^3$INESC-ID and Instituto Superior T\'ecnico, Universidade de Lisboa, Portugal \\
		$^4$ Machine Learning Group, Universit{\'e} Libre de Bruxelles, 1050 Brussels, Belgium  \\ 
		$^5$ Artificial Intelligence Lab, Vrije Universiteit Brussel, 1050 Brussels, Belgium \\
		$^6$ Center for Human-Compatible AI, UC Berkeley, Berkeley, 94702, USA \\
		$^7$ FARI Institute, Universit{\'e} Libre de Bruxelles-Vrije Universiteit Brussel, 1050 Brussels, Belgium \\
		$^\star$ Corresponding author: The Anh Han (T.Han@tees.ac.uk)
	}

\newpage
\section*{Abstract}
Regulation of advanced technologies such as Artificial Intelligence (AI)  has become increasingly {important, given  the associated risks and apparent }ethical issues. With the great benefits promised from being able to first supply  such technologies, safety precautions and societal consequences might be ignored or shortchanged in exchange for speeding up the development, therefore engendering a racing narrative among the developers. Starting from a game-theoretical model describing an idealised technology race in a {fully connected} world of players, here we investigate how different interaction structures among race participants can alter collective choices and requirements for regulatory actions. Our findings indicate that, when participants portray a strong diversity in terms of connections and peer-influence (e.g., when scale-free networks shape interactions among parties), the conflicts that exist in homogeneous settings are significantly reduced, thereby lessening the need for regulatory actions. Furthermore, our results suggest that technology governance and regulation may profit from the world's patent heterogeneity and inequality among firms and nations, so as to enable the design and implementation of meticulous interventions on a minority of participants, which is capable of influencing an entire population towards an ethical and sustainable use of advanced technologies. \\
 
 \noindent \textbf{Keywords:} AI Safety, Complex Networks, Evolutionary Game Theory, Agent-based Simulation. 

 \newpage
 

\section{Introduction}
Researchers and stakeholders alike have urged for due diligence in regard to AI development on the basis of several {concerns}. Not least among them is that AI systems could easily be applied to {nefarious purposes}, such as espionage or cyberterrorism \citep{taddeo2018regulate}. Moreover, the desire to be at the foreground of the state-of-the-art or the pressure imposed by upper management might tempt developers to ignore {safety procedures} or ethical consequences \citep{armstrong2016racing,cave2018ai}. 
Indeed, {such} concerns have been expressed in {many forms, from letters} of scientists against the use of AI in military applications \citep{FLI_letter2015,FLI_signatories_todate}, {to} blogs of AI experts requesting careful communications \citep{BrookBlog2017}, {and proclamations} on {the} ethical use of AI \citep{MontrealDec2018,steels2018barcelona,russell2015ethics,jobin2019global}. 


Regulation and governance of advanced technologies such as Artificial Intelligence (AI)  has become increasingly more important given their potential implications, such as  associated risks  and ethical issues \citep{EUAIwhitepaper2020,MontrealDec2018,steels2018barcelona,russell2015ethics,jobin2019global,FLI_letter2015,FLI_signatories_todate,perc2019socialAI}. With the great benefits promised from being first able to supply  such technologies, stake-holders might cut corners on safety precautions in order to ensure a rapid deployment{, in a race towards AI market supremacy (AIS)} \citep{armstrong2016racing,cave2018ai}.  
One does not need to look very far to find {potentially} disastrous scenarios associated with AI \citep{sotala2014responses, armstrong2016racing, pamlin2015global,o2016weapons}, but accurately predicting outcomes and accounting for these risks is exceedingly
difficult in the face of uncertainty \citep{armstrong2014errors}.
As part of the double-bind problem put forward by the Collingridge Dilemma, the impact of a new technology is difficult to predict {before it has been already extensively developed and widely adopted, and also difficult to control or change after it has become entrenched} \citep{Collingridge1980}. Given the lack of available data and the inherent unpredictability involved in {this new field of technology}, a modelling approach is therefore desirable to provide a better grasp of any expectations with regard to a race for AIS. Such modelling allows for dynamic descriptions of several key features of the AI race (or its parts), providing an understanding of possible outcomes, considering external factors and conditions, and the ramifications of any policies that aim to regulate such race.

With this aim in mind, a baseline model of an innovation race has been recently proposed \citep{han2019modelling}, {in which innovation dynamics are pictured through the lens of Evolutionary Game Theory (EGT) and all race participants are equally well-connected in the system (well-mixed populations). The baseline results showed the importance of accounting for different time-scales of development, and also exposed the dilemmas that arise when what is individually preferred by developers differs from what is globally beneficial. When domain supremacy could be achieved in the short-term, unsafe development required culling for to promote the welfare of society, and the opposite was true for the very long term{, to prevent excessive regulation at the start of exploration}. }
However, real-world stakeholders and their interactions are far from homogeneous. Some individuals are more influential than others, or play different roles in the {unfolding} of new technologies. Technology races are shaped by complex networks of exchange, influence, and competition where diversity abounds. 
It has been shown that particular {networks} of contacts can promote the evolution of positive behaviours in various settings, including cooperation  \citep{santos2006evolutionary,ohtsuki2006simple, santos:2008:nature,perc2017statistical,chen2015first,perc2010coevolutionary}, fairness \citep{page2000spatial,szolnoki2012defense,wu2013adaptive,santos2017structural,cimpeanu2021cost} and trust \citep{Kumar_2020}.  In this paper, {we take inspiration from the disconnect between the previous line of research and the heterogeneity observed in real-world interactions}, and ask whether network topology can influence the adoption of safety measures in innovation dynamics, and shape {the tensions of the AI development race}.

{The impact of network topology } is particularly important in the context of technology regulation and governance. {Technology} innovation and collaboration networks (e.g. among firms, stakeholders and AI researchers)  are highly heterogeneous \citep{schilling2007interfirm,newman2004coauthorship}. 
Developers or development teams interact more frequently within their groups {than without}, forming alliances and networks of followers and collaborators \citep{barabasi2014linked,ahuja2000collaboration}. Many companies  compete in several markets while others compete in only a few, and their positions in inter-organisational networks strongly influence their behaviour (such as resource sharing) and innovation outcome   \citep{ahuja2000collaboration,shipilov2020integrating}. 
{It is important to understand how diversity in the network of contacts
influences race dynamics and the conditions under which regulatory actions are needed}. Therefore, {we} depart from a minimal AI race  model \citep{han2019modelling}, examining {instead} how network structures  influence safety decision making within an AI development race. 
 
In a structured population, players are competing with {co-players in their {network} neighbourhoods}. Firms interact or directly compete through complex ties of competition, such that some players may play a pivotal role in a global outcome. Here we abstract these relationships as a graph or a network. We compare different forms of network structures, from homogeneous ones --- such as {complete} graphs {(equivalent to well-mixed populations),} and square lattices --- to different types of scale-free networks \citep{barabasi1999emergence} ({see Methods}), representing different levels of diversity in the number of co-player races a player can compete in. Our results show that when race participants are distributed in a  heterogeneous network, the conflicting tensions arising in the well-mixed case are significantly reduced, thereby softening the need for regulatory actions. 
This is, however, not the case when the network is not accompanied by some degree of relational heterogeneity, even in different types of spatial lattice networks. 

{In the following sections, we describe the models in detail, then present our results.}


\section{{Models and Summary of Previous Results}} 
\label{section:models and methods}
{We first define the AI race game \citep{han2019modelling}  and recall relevant results from previous works in the well-mixed populations setting.  }  
\subsection{AI race model definition}
{
Assuming that winning the race towards supremacy is the goal of the development teams (or players) and that a number of development steps (or advancements/rounds) are required, the players have two strategic options in each step: to follow safety precautions (denoted by  strategy SAFE) or to ignore them (denoted by { strategy} UNSAFE) \citep{han2019modelling}. As it takes more time and effort to comply with the precautionary requirements, playing SAFE is not only costlier, but also implies a slower development speed, compared to playing UNSAFE. Let us also assume that to play SAFE, players need to pay a cost $c$, whereas the opposite { strategy}  is free. The increase in speed when playing UNSAFE is given by a free parameter $s > 1$, while the speed when playing SAFE is normalised to 1. The interactions are iterated until one or more teams achieve a designated objective, after having completed $W$ development steps. As a result, the players obtain a large benefit $B$, shared among those who reach the target objective at the same time. However, a setback or disaster can happen with some probability, which is assumed to increase with the number of times  the safety requirements have been omitted by the winning team(s). Although many potential AI disaster scenarios have been sketched \citep{armstrong2016racing,pamlin2015global}, the uncertainties in accurately predicting these outcomes  are high.
 When such a disaster occurs, risk-taking participants lose all their benefits.  We denote by $p_r$ the risk probability of such a disaster occurring when no safety precaution is followed at all.}

We model {an} AI development race as a repeated two-player game, consisting of $W$ development rounds. In each round, the players can collect benefits from their intermediate AI products, depending on whether they choose to play SAFE or UNSAFE. Assuming a fixed benefit $b$,  from the AI market, {teams share} this benefit proportionally to their development speed. Moreover, we assume that with some probability $p_{\textit{fo}}$ those playing UNSAFE might be found out, wherein their disregard for safety precautions is exposed, leading to their products not being adopted {due to safety concerns}, thus receiving 0 benefit.  
Thus, in each round of the race, we can write the payoff matrix as follows (with respect to the row player) 
\begin{equation}
	\label{payoffs-simple}
\Pi =  \bordermatrix{~ & \textit{SAFE} & \textit{UNSAFE}\cr
                  \textit{SAFE} & -c + \frac{b}{2} &-c + (1-p_{\textit{fo}}) \frac{b}{s+1}    + p_{\mathit{fo}} b  \cr
                  \textit{UNSAFE} & (1-p_{\textit{fo}}) \frac{s b}{s+1}   & (1-p^2_{\textit{fo}}) \frac{ b}{2}   \cr
                 }.
\end{equation}

For instance, when two SAFE players interact, each needs to pay the cost $c$ and they share the benefit $b$. 
When a SAFE player interacts with an UNSAFE one, the SAFE player pays a cost $c$ and  obtains {(with probability $p_\mathit{fo}$)} the full benefit $b$ 
in case the UNSAFE co-player is  found out, and obtains {(with probability $1 - p_\mathit{fo}$)} a small part of the benefit $b/(s+1)$ otherwise, dependent on the co-player's speed of development $s$.  When playing with a SAFE player,  the UNSAFE one does not have to pay any cost and obtains 
a larger share $bs/(s+1)$ when not  found out. Finally, when an UNSAFE player interacts with another  one, it 
obtains the shared benefit $b/2$ when both are not found out{, {but }the full benefit ${b}$} when it is not found out while the co-player is found out, and 0 otherwise. The corresponding {average} payoff is thus:  $(1-p_\mathit{fo})\left[  (1-p_\mathit{fo}) (b/2) + p_\mathit{fo} b \right] = (1-p^2_\mathit{fo}) \frac{ b}{2}.$ 

In the AI development process, players repeatedly interact (or compete) with each other using the \textit{innovation} game described above. In order to clearly examine the effect of population structures on the overall outcomes of the AI race, in line with previous network reciprocity analyses (e.g. in social dilemma games \citep{santos2006pnas,santos:2008:nature,Szabo2007}), we focus in this paper on two unconditional strategies \citep{han2019modelling}:
\begin{itemize} 
	\item AS (always complies with safety precautions)  
	\item AU (never  complies with safety precaution{s})  
\end{itemize} 
Denoting by $\Pi_{ij}$ ($i, j \in \{1,2\}$) the entries of the matrix $\Pi$ above, the payoff matrix defining the averaged payoffs for {AU vs AS} reads   
\begin{equation}
	\label{payoff-average}
 \bordermatrix{~ & \textit{AS} & \textit{AU} \cr
                  \textit{AS} & \frac{B}{2W} +\Pi_{11} & \Pi_{12}    \cr
                  \textit{AU} &  (1 - p_r) \left(\frac{sB}{W} + \Pi_{21}\right)   &  (1 - p_r) \left(\frac{sB}{2W} +\Pi_{22}\right)  \cr 
                 }.
\end{equation}

As described in Equation \ref{payoff-average}, we encounter the following scenarios. When {only} two safe players interact, they complete the race simultaneously after an average of $W$ development rounds, thereby obtaining the averaged split of the full prize $\frac{B}{2W}$ per round; furthermore, the safe players also obtain the intermediate benefit per round ($\pi_{11}$, see Equation \ref{payoffs-simple}). When a safe player {only} encounters an unsafe player, the only benefit obtained by the safe player is the intermediary benefit in each round, {whereas the unsafe player receives the full prize $B$}; {moreover, }the unsafe player completes the race in $\frac{W}{s}$ development rounds, so it receives an {extra} average of $\frac{sB}{W}$ of the full prize per round. Furthermore, the unsafe behaviour attracts the possibility of {a disaster occurring, causing them to lose all gains, with probability $p_r$, which is reflected in the payoff matrix {(consider $\pi_{22}$ in Equation \ref{payoffs-simple})}}. Similarly, we can extract the {average} payoffs for {solely} two unsafe players interacting, by considering that they finish the race at the same time and get the appropriate intermediate benefit $\pi_{22}$ (See Equation \ref{payoffs-simple}).

\subsection{{Summary of previous results in well-mixed settings}}
{
In order to clearly present the contribution of the present work, we next  recall the analytical conditions derived in \citep{han2019modelling} and how these will be used to inform the analysis that follows. Our analysis will  differentiate  between two development regimes: an early/short-term regime and a late/long-term one}. The difference in time-scale between the two regimes plays a key role in identifying which regulatory actions are needed and when. {This distinction is in line with previous works adopting analytical approaches using stochastic population dynamics \citep{han2019modelling}.}
The early regime is underpinned by {the race participants' ability to readily reach the ultimate prize \textit{B}} in the shortest time frame available. In other words, winning the ultimate prize in \textit{W} rounds is much more important than any benefits achieved in single rounds until then, i.e. $B/W >> b$. Contrarily, a late regime is defined by a desire to do well in {each development round}, as {technological supremacy cannot} be achieved in the foreseeable future. {That is}, singular gains \textit{b}, even when accounting for the safety cost \textit{c}, become more tempting than aiming towards winning the ultimate prize, i.e. $B/W << b$. For a reminder of the meanings of the parameters described above, see Table \ref*{table:parameters}.

\begin{table}[]
\caption{Model parameters and parameter space analysed}
\label{table:parameters}
\begingroup
\renewcommand{\arraystretch}{1.25}
\begin{tabular}{lcc}
\hline
 Parameter  &  Symbol & Range Analysed\\ \hline
 Population size & $Z$ & \{100, 1000, 1024\} \\
 Intensity of selection & $\beta$  & \{$1$\}  \\
 Average connectivity of a scale-free network & $z$  & \{$4$\}  \\
 Number of new edges for each new node in SF networks & $m$  & \{$2$\}  \\
 Probability of being found out when playing unsafe & $p_{fo}$  & \{0, 0.05, 0.1, ..., 1\}  \\
 Probability of disaster occurring due to unsafe development & $p_{r}$  & \{0, 0.05, 0.1, ..., 1\}  \\
 Benefit of winning the race (reaching AI supremacy) & $B$  & \{$10^4$\}  \\
 Benefit of intermediate AI advancements  & $b$  & \{$4$\}  \\
 Cost of adhering to safety standards & $c$  & \{$1$\}  \\
 Speed of development (due to disregarding safety) & $s$  & \{1, 1.25, 1.5, ..., 5\}  \\
 Number of development rounds until AI supremacy is reached & $W$  & \{$100, 10^6$\}  \\ \hline
\end{tabular}
\endgroup
\end{table}

\begin{figure}
	\centering
	\includegraphics[width=1.0\linewidth]{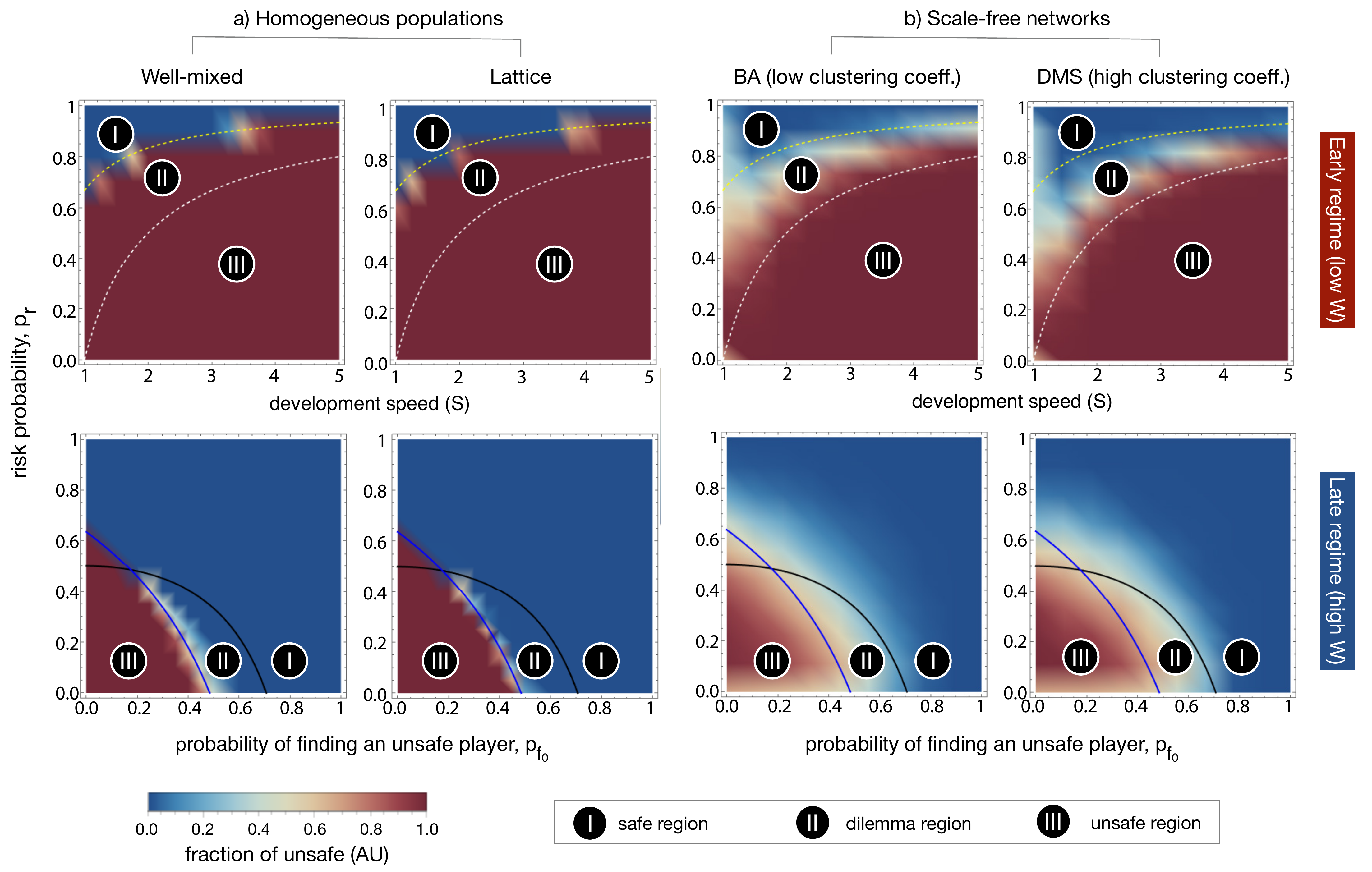}
	\caption{Color gradients indicating the average fraction of AU (unsafe strategy) for (a) homogeneous (well-mixed and lattices) populations and (b) scale-free networks (BA and DMS models). The top row addresses the early regime (low $W$) for varying development speed ($s$) and risk probability ($p_r$). The bottom row addresses the late regime (high $W$) for varying $p_{fo}$ (the chances that an UNSAFE player is found out) and risk probability ($p_r$). Dotted and full lines indicate the phase diagram obtained analytically \citep{han2019modelling}. In the early regime (upper panels), region II indicates the parameters in which safe AI development is the preferred collective outcome, but unsafe development is expected to emerge and regulation may be needed --- thus the dilemma. In regions I and III, safe and unsafe AI development, respectively, are both the preferred collective outcomes and the ones expected to emerge from self-organization, hence not requiring regulation. In the late regime (lower panels), the solid black line marks the boundary above which safety is the preferred outcome, whereas the blue line indicates the boundary above which safety becomes risk dominant against unsafe development. The results obtained for well-mixed populations and lattices (a) suggest that, for both early and late regimes, the nature of the dilemma, as represented by the analytical phase diagram, remains unaltered. Moreover, homogeneous interaction structures cannot reduce the need for regulation in the early regime. Differently, we show that heterogeneous interaction structures (scale-free networks, (b)) are able to significantly reduce the prevalence of unsafe behaviors for almost all parameter regions, including both late and early regimes. This effect is enhanced whenever scale-free networks are combined with high clustering coefficient (i.e., in the DMS model). Other parameters: $p_{fo}=0.5$, and $W=100$ (top panels); $s = 1.5$ and  $W = 10^6$ (bottom panels); $c = 1$, $b=4$, $B=10^4$, and $\beta=1$, in all panels.}
	\label{fig:allnets}
\end{figure}

We {have also made use} of the previous analytical results \citep{han2019modelling} which identify the risk-dominant boundaries of the AI race game for both early and late development  regimes in well-mixed populations. {These} are useful as a baseline or reference model, determining the  regions in which regulatory actions are needed or otherwise, and moreover, if needed, which behaviour should be promoted. 
In the early regime, the two dotted lines mark {region (II)} within the boundaries $p_r \in [1-1/s, \ 1-1/(3s)]$ for which safety development is the preferred collective outcome, but where unsafe development is selected for {by social dynamics} (see e.g. Figure  \ref{fig:allnets}, first row). {Thus, in this region (II), regulation is required to improve safety compliance.}  Outside of these boundaries, safe {(in region I) and unsafe (in region III)}, respectively, are both the preferred collective outcomes and the ones selected for by social dynamics{, hence requiring no regulatory actions}. For the late AI race (e.g. Figure  \ref{fig:allnets}, bottom row), the solid black line marks the boundary above which safety is the preferred collective outcome, where $p_r < 1 - \frac{b - 2c}{b(1 - p_{fo}^2)}$, whereas the blue line indicates where AS becomes risk-dominant against AU, where $p_r < \frac{4c(s + 1) + 2b(s - 1)}{b(1 + 3s)}$. Again, in this regime three regions can be distinguished, with (I) and (III) having similar meanings to those in the early regime. However, differently from the early regime, in region (II) regulatory actions are needed to improve (unsafe) innovation instead of safety compliance, due to the low risk. These regions are derived from the analytical conditions described in \citep{han2019modelling}, where these are explained in further detail.

\section{Results} 
{Based on extensive computer simulations (see Methods), our analysis identifies the prevalence of individuals adopting unsafe procedures after reaching a stationary state and infer the most likely behavioural trends and patterns} associated with the agents taking part in the AI race game for distinct network topologies. {The main findings from this work are described in this section, whereby each subsection will provide a key insight followed by the results and intuitions which motivate each claim.}

\subsection{Heterogeneous interactions reduce the viability of unsafe development in both short and long-term races}

We examine the impact of different network structures, homogeneous and heterogeneous, on the safety outcome of the evolutionary dynamics for the two different development regimes descried above.  To commence our analysis, we {first} study the role of degree-homogeneous graphs {(here illustrated by structural spatiality) in the evolution of strategies in the AI race game}. {First}, we simulated the AI race game in well-mixed populations (see Figure \ref{fig:allnets}, first column). We then explored the same game on a square lattice, where each agent can interact with its four edge neighbours, in Figure \ref{fig:allnets} (second column). We show that the trends remain the same when compared with well-mixed populations, with very slight differences in numerical values between the two. Specifically, towards the top of area (Region \textbf{II}), at the risk-dominant boundary between AS and AU players in the case of an early AI race, we see some safe developmental activity where previously there was none. In practice, this shifts the boundary very slightly towards an optimal conclusion.

Thus, except for minute atypical situations, we may argue that homogeneous spatial variation is not enough to influence safe technological development, with minimal improvement when compared with a well-mixed population (complete network). To further increase our confidence that such structures have very small effects on the AI race game, we confirm that 8-neighbour lattices (where agents can also interact with corner neighbours) yield very similar trends, with negligible differences when compared to either the regular square lattice or well-mixed populations (see Supplementary Information, Figure S2).


As a means of investigating beyond simple homogeneous structures and their roles in the evolution of appropriate developmental practices in the AI race, we make use of the previously defined BA and DMS network models (see {Methods}). Contrary to the findings on homogeneous networks, scale-free interaction structures produce marked improvements in almost all parameter regions of the AI race game (see Figure \ref{fig:allnets}).  


\begin{figure}
	\centering
	\includegraphics[width=0.9\linewidth]{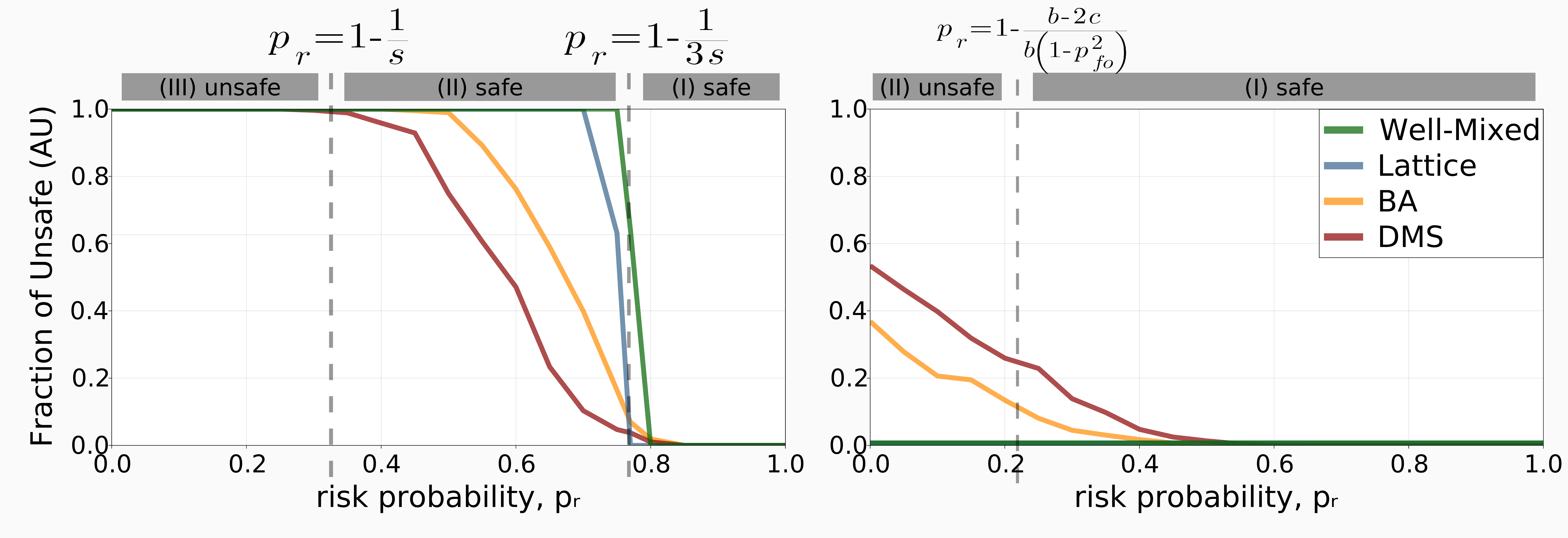}
	\caption{Heterogeneous networks moderate the need for regulation, shown by measuring frequency of unsafe developments  across a range of different risk probabilities. The boundaries between zones are indicated with  blue dashed lines, whereas the grey-highlighted texts on top of the figures indicate the collectively desired behaviour in each zone. The left panel reports the results for the early regime ($p_{fo} = 0.5, \ W = 100$), while the right panel does so for the late regime ($p_{fo} = 0.6, \ W = 10^6$) (parameter values are chosen for a clear illustration).  Parameters: $c = 1, \ b = 4, \ B = 10^4, \ s = 1.5, \  \beta = 1.$}
	\label{fig:dilemma-zones}
\end{figure}

Previously, it has been suggested that different approaches to regulation were required, subject to the time-line and risk region in which the AI development race is placed, after inferring the preferences developers would have towards safety compliance \citep{han2019modelling}. Given that innovation in the field of AI (or more broadly, technological advancement as a whole), should be profitable (and robust) to developers, shareholders and society altogether, we must therefore discuss the analytical loci where these objectives can be fulfilled. Assuredly, we see that diversity in players introduces two marked improvements in both early and late safety regimes. Firstly and most importantly, we note that very little regulation is required in the case of a late AI race (large $W$), principally concerning the existing observations in homogeneous settings (e.g., well-mixed populations and lattices). Intuitively, this suggests that there is little encouragement needed to promote risk-taking in late AIS regimes: Diversity enables {beneficial innovation}. Secondly, the region for early AIS regimes in which regulation must be enforced is diminished, but not completely eliminated. {Consequently}, governance should still be prescribed when developers are racing towards an early or otherwise unidentified AI regime (based on the number of development steps or risk of disaster). It stands to reason that insight into what regime type the AI race operates in, is therefore paramount to the success of any potential regulatory actions. The following sections will attempt to look further into assessing these observations.

Figure \ref{fig:allnets} (top panels) presents a fine-grained glimpse into the early regime. In region (\textbf{II}), the safety dilemma zone, social welfare is once more conspicuously improved  by heterogeneity. {Concerted} safe behaviour is favoured, even in the face of being disregarded by social dynamics in the analytical sense. We discern the clear improvements discussed earlier, but also echo the messages put forward in \citep{han2019modelling}. We contend that it is vital for regulators to intervene {in these conditions}, for encouraging pro-social, safe conduct, and in doing so avert conceivably dangerous outcomes. {Heterogeneity} lessens the burden on policy makers, allowing for greater freedom in the errors and oversights that could occur in governing towards the goal of safe AI development.

While the difference between heterogeneous and homogeneous networks is evident, there also exists a distinction between the different types of heterogeneous networks. In this paper we discuss the BA and DMS models, and also their normalised counterparts, in which individuals' payoffs are divided by the number of neighbours. In such scenarios one could assume that there is an inherent cost to maintaining a link to another agent. In this sense, there exists some levelling of the payoffs, {seemingly} increasing fairness and reducing wealth inequality. But we confirm that normalising the network leads to similar dynamics {as observed in} homogeneous populations (see Figure S3), with only very {slight differences}. 

In order to accurately depict the measured differences between the different types of networks, we varied the risk probability ($p_r$) for both the early and  late regime. We report the results of this analysis in Figure \ref{fig:dilemma-zones}, where we also show the preferred collective outcome, using the different regions described {earlier in} this section. These figures help expose the effect of heterogeneity on the frequency of unsafe behaviour in the different dilemma zones. In particular, we notice a {mediating effect in the requirements for regulation,} for both regimes and types of scale-free networks. 

Specifically, in the case of the early regime (see Figure \ref{fig:dilemma-zones}, left column), we observe the presence of safety for a much broader range of risk probability values, than in the case of either well-mixed or structured populations. In the late regime (see Figure \ref{fig:dilemma-zones}, right column), however, we also highlight an increase in unsafe behaviour even beyond the boundary for which safety would have been the preferred collective outcome. In this case, heterogeneity has its drawbacks. On the one hand, innovative behaviour sees some improvement when it is in the interest of the common good for it to be so, but the same is true, albeit rarely, when it is not. We also note that the effects described above are amplified in the case of DMS networks, in comparison to their BA counterparts. {Observing a high degree of inter-group interactions (clustering)} may play a key role in determining if intervention is required in the AI race. Moreover, we confirm these findings by producing typical runs showing the time evolution of unsafe behaviour for each network type (please see Figure S1).


\begin{figure}
	\centering
	\includegraphics[width=0.9\linewidth]{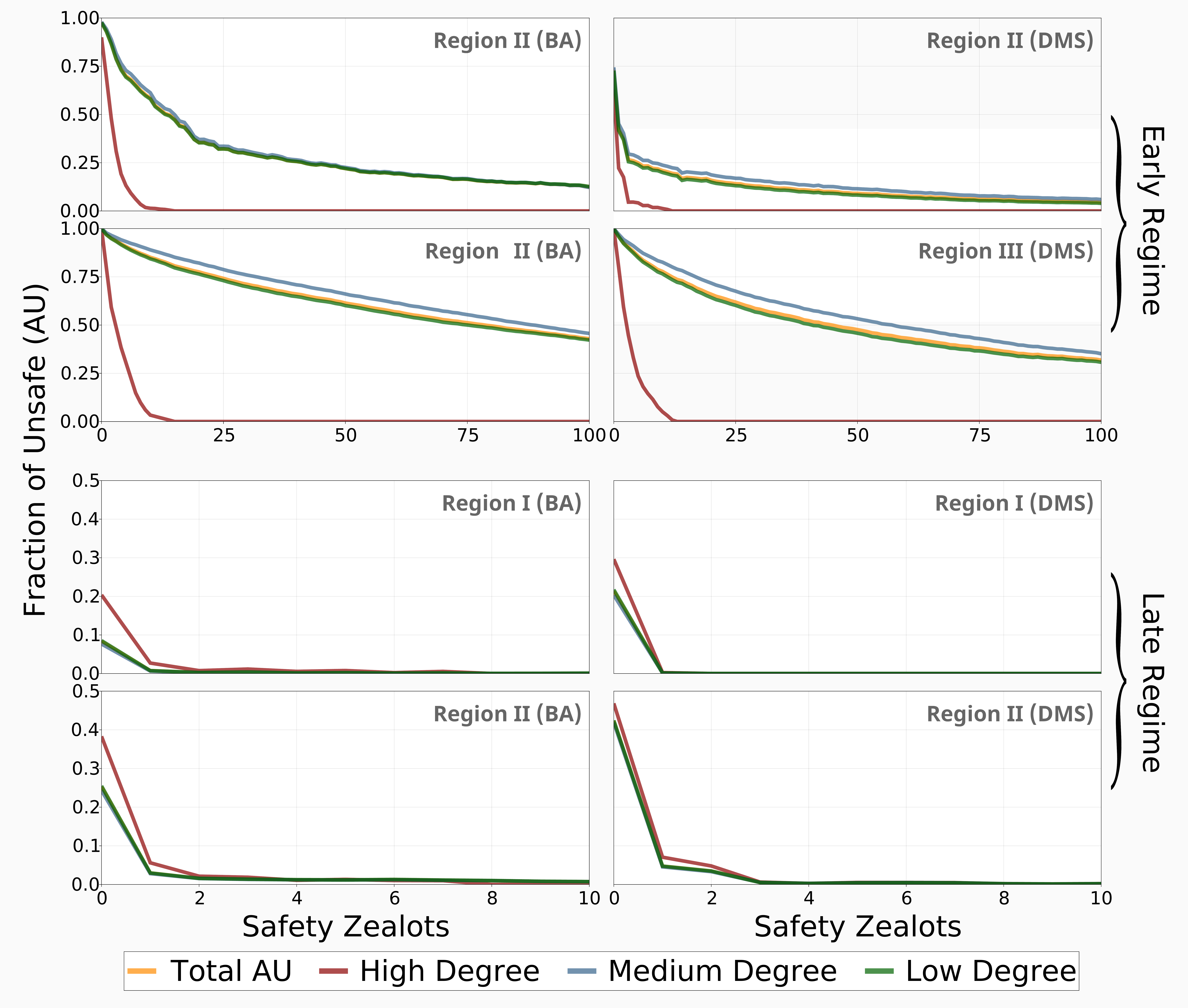}
	\caption{Hubs prefer slower, thus safer developments in the early race, and this can be further exploited by progressively introducing safety zealots in highly connected nodes. We show the results for both regimes, as well as the appropriate regions where safety (early region II and late region I), and conversely where innovation (early region III and late region II) are the preferred collective outcomes. The top four panels report the results for the early regime ($p_{fo} = 0.5, \ W = 100$ with $p_r = 0.5$ for region II and $p_r = 0.1$ for region III), and the bottom four do so for the late regime ($p_{fo} = 0.6, \ W = 10^6$ with $p_r = 0.3$ for region I and $p_r = 0.1$ and region II). We show a subset of the results in the late regime for clear representation; see Figure S8 for a comprehensive view.  Other parameters: $c = 1, \ b = 4, \ B = 10^4, \ s = 1.5, \  \beta = 1.$}
	\label{fig:zealots-scale-free}
\end{figure}

\subsection{Hubs and their role in decelerating the race}

Highly connected individuals (hubs) typically play a key role in many real-world networks of contacts and {change the dynamics observed in} heterogeneous populations \citep{pastor2001epidemic,santos2006pnas,santos:2008:nature,perc2010coevolutionary}. In order to study the role {that} hubs play in the AI race, in the context of scale-free networks, we classify nodes into three separate connectivity classes \citep{santos:2008:nature}. We obtain three classes of individuals, based on their number of contacts (links) $k_i$ and the average network connectivity $z$: 

\begin{enumerate}
	\item Low degree, whenever $k_i < z$,
	\item Medium degree, whenever $z \leq k_i < \frac{k_{max}}{3} $ and
	\item High degree (hubs), whenever $\frac{k_{max}}{3} \leq k_i \leq k_{max}$.
\end{enumerate}

Dedicated minorities are often identified as major drivers in the emergence of collective behaviours in social, physical and biological systems, see \citep{pacheco2011messianic,santos2019evolution,paiva2018engineering,Cardillo_2020}. Given the relative importance of hubs in other systems, we explore whether highly connected, committed individuals are prime targets for safety regulation in the AI race. By introducing individuals with {pathologically safe tendencies (fixed behaviours)} \citep{santos2019evolution}---these are sometimes referred to as zealots, see \citep{pacheco2011messianic, santos2019evolution, Cardillo_2020, Kumar_2020}---in the network, we can better understand the power of influential devotees in the safe development of a general AI. 

We progressively introduce pathological safe players based on {their} degree centrality (i.e. number of connections). {In other words, the most connected nodes will be the first to be targeted.} The benefits of this approach are twofold, as they allow us to study the relative differences between the three classes of individuals, but also the effect of regulating the key developers in the AI race. For a full analysis of the differences between high, medium and low degree individuals in the baseline case, please see Figure S5.

Hubs prefer slower, safer developments in the early AI race, and this can be further exploited by introducing safety zealots in key locations in the network (see Figure \ref{fig:zealots-scale-free}). When safety is the preferred collective outcome, hubs can drive the population away from unsafe development, and this effect is even more apparent in the case of highly clustered scale-free networks (Figure \ref{fig:zealots-scale-free}, right column). Following the sharp increase in global safety after the conversion of high degree players to zealotry, we also observe a similar, but not as pronounced influence as the most connected medium degree individuals follow suit, an effect which plateaus shortly thereafter. We further confirm these results by selecting the same targets (the top $10\%$ of individuals based on degree centrality), but introducing them in reverse order (i.e. starting with the highest connected medium degree individuals and ending with the most connected high degree ones; see Figure S9). 

Whereas the capacity of hubs to drive the population towards safety is evident in region II of the early regime (when safety is the {collective} preferred outcome), the opposite is true for region III. High degree individuals are more capable at influencing the overall population than medium degree individuals{, even the most highly connected ones}, but we see a much more gradual decrease in innovation as the most connected nodes are {steadily} converted to zealotry. Even in the presence of great uncertainty, a small percentage of very well connected developers can ensure safety with very little negative impact on innovation. 

Conversely, we show that highly connected individuals prefer innovation in the late regime, irrespective of the preferred collective outcome. {But even the introduction of one pathological safe player ($0.1\%$ of the population) in the largest hub} is enough to ensure that the entire population converges to safe development in most instances. In the case when safety is {socially} preferred (third row of Figure \ref{fig:zealots-scale-free}), the successful regulation of the AI race requires a very small minority of individuals to dedicate themselves to safety, but in cases of uncertainty, innovation is very easy to stifle in the late regime, even  when it would be beneficial not to do so (region II).


 
\begin{figure}
	\centering
	\includegraphics[width=0.9\linewidth]{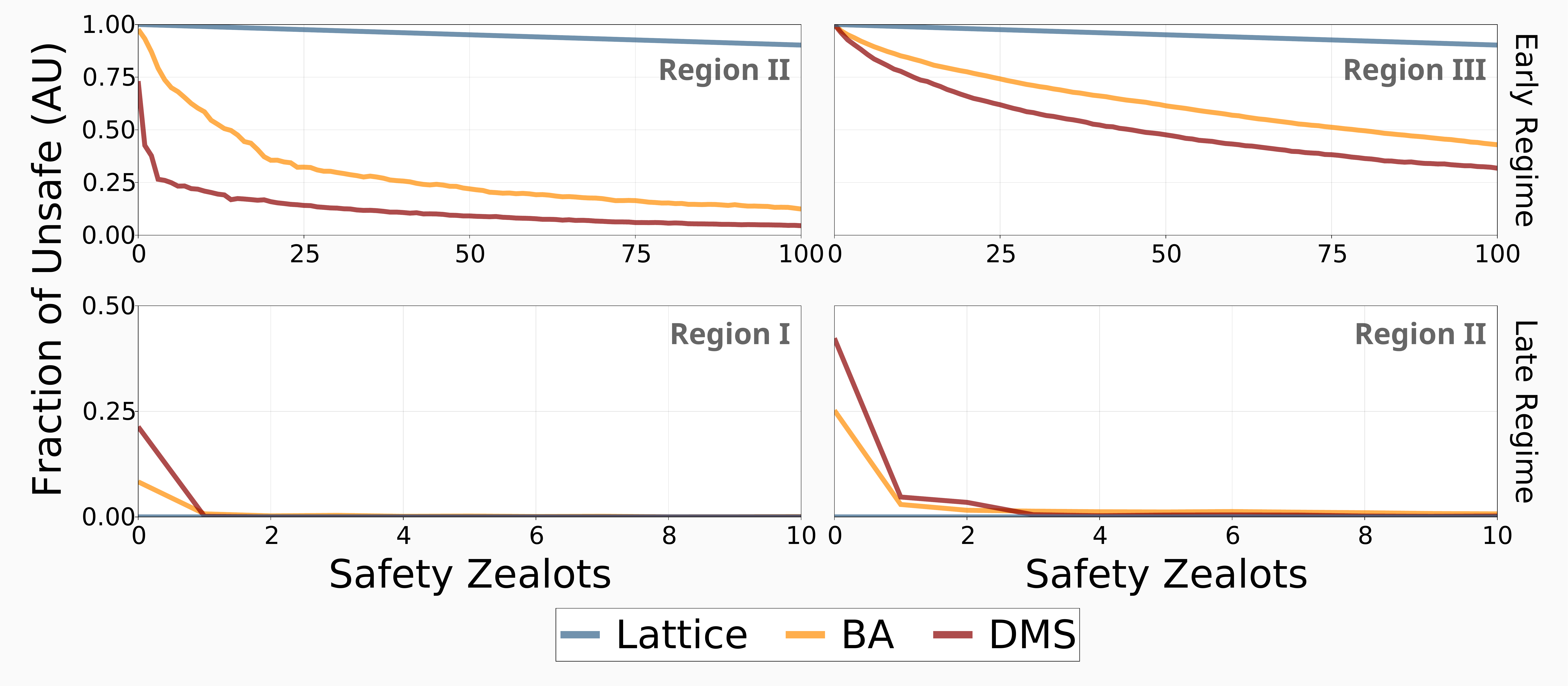}
	\caption{Introducing a small number of safety zealots can mitigate race tensions under uncertainty. We show the results for both regimes, as well as the appropriate regions where safety (early region II and late region I), and conversely where innovation (early region III and late region II) are the preferred collective outcomes. The top panels report the results for the early regime ($p_{fo} = 0.5, \ W = 100$ with $p_r = 0.5$ for region II and $p_r = 0.1$ for region III), and the bottom do so for the late regime ($p_{fo} = 0.6, \ W = 10^6$ with $p_r = 0.3$ for region I and $p_r = 0.1$ and region II). We note that these values were chosen for clear representation.  Other parameters: $c = 1, \ b = 4, \ B = 10^4, \ s = 1.5, \  \beta = 1.$}
	\label{fig:zealots-networks}
\end{figure}

\subsection{A small minority of highly connected individuals can help mitigate race tensions under uncertainty}
\label{zealots}

Uncertainty can limit the options of regulatory agencies in the quest towards the development of safe AI, and narrow solutions to regulation could have potentially disastrous consequences, given the existential risk that general AI poses to humanity \citep{Scherer2015}. Moreover, the promised benefit of such technology is great enough that stifling innovation could be nearly as harmful as the catastrophic consequences themselves, given the potential solutions that such {technology} could provide to problems {in the context of} existential risk, healthcare, politics, and many other fields {(e.g. \citep{McKinney2020, rolnick2019tackling})}. 

To provide general solutions to the problem of regulating the AI race, we explore the impact of safety zealots (as discussed in the previous section) across the whole range of possible scenarios. {We} cannot be sure of the nature of the network of contacts that governs real-world AI developers, nor the actual timeline of the race. We show that by enforcing safety for a very select minority of highly connected individuals, race tensions can be mitigated in nearly all cases (see Figure \ref{fig:zealots-networks}). We provide a full analysis of the effect of zealots in well-mixed networks in Figure S4, and note that the lack of heterogeneity produces nearly identical results to lattice networks. 

Slowing key individuals in the early regime can dramatically reduce existential risk in the case of heterogeneous interactions. For both regions, hubs in DMS networks can drive the other nodes towards safety (see Figure \ref{fig:zealots-networks}, top panels), but the reduction in unsafe developments in region II is significantly higher than in region III for low numbers of safety zealots. Outside of the few individuals that are converted to zealots, other nodes maintain their speed and continually innovate in region III, which suggests that this approach could be fundamental to the governance of developmental races. We note that if the proportion of safety zealots is not high enough, this effect cannot be reproduced, even in the presence of additional interference (such as artificially funding zealots or accelerating their development); For a more thorough analysis, please see Figures S6 and S7.

Given a drawn-out race, this small minority of zealots can negatively impact innovation in the late regime (region II), where the relative increase that heterogeneous interactions provided rapidly disappears as pathological players are introduced. On the other hand, conditional strategies have been shown to further diminish the need to promote innovation in these conditions \citep{han2019modelling}, and the introduction of these advanced strategies in this model could eliminate the negative effects of safety zealots in this region.

\section{Discussion}

{In this paper}, we have considered the implications of network dynamics on a technological race for supremacy in the field of AI, with its implied risks and hazardous consequences \citep{sotala2014responses,armstrong2016racing,pamlin2015global}. We make use of a previously proposed evolutionary game theoretic model \citep{han2019modelling} and study how the tension and  temptation resulting from the  race can can be mediated, for both early and late development regimes. 

Network reciprocity has been shown to promote the evolution of various positive outcomes in many settings \citep{santos:2008:nature,ranjbar2014evolution, szolnoki2012defense, wu2013adaptive} and, given the high levels of heterogeneity identified in the networks of firms, stakeholders and AI researchers \citep{schilling2007interfirm,newman2004coauthorship}, it is important to understand the effects of reciprocity and how it shapes the dynamics and global outcome of the development race. It is just as important to ensure that appropriate context-dependent regulatory actions are provided. {This modelling approach and the associated results are} applicable to other technologies and competitions, {such as patent races or the development of biotechnology, pharmaceuticals, and climate change mitigation technology, where there is a significant advantage to be achieved by reaching some target first \citep{denicolo2010winner,campart2014technological,lemley2012myth,abbott2009global,burrell2020covid}. Given a sufficiently tempting potential gain, individuals are more likely to invest in high-risk technology \citep{Andrews2018}, which suggests that these insights could be applicable to many similar fields in which risk and innovation must be constantly balanced.}

It is noteworthy that, despite  a number of proposals and debates on how to prevent, regulate, or resolve an AI race \citep{baum2017promotion,cave2018ai,geist2016s,shulman2009arms,vinuesa2019role,taddeo2018regulate,askell2019role,hanAIES2019}, only a few formal modelling studies have been  proposed  \citep{armstrong2016racing,han2019modelling,han2020Incentive,han2021voluntary}. These works focus on homogeneous populations, where there are no inherent structures indicating the network of contacts among developing teams.
{Innovation dynamics (including AI) emerge from complex systems marked by a strong diversity in influence and companies' power. Firms create intricate networks of concurrent development, in which some develop a higher number of products, influencing and competing with a significant number of others. Our work} advances this line of research, revealing the impact of these network structures among race participants, on the dynamics and global outcome of the development race.

We began by validating the analytical results obtained as a baseline in a completely homogeneous population \citep{han2019modelling}, using extensive agent-based simulations. We then adopted a similar methodology to analyse the effects of gradually increasing network heterogeneity, equivalent to  diversifying the connectivity and influence of the race participants. {This was accomplished by studying square lattices,} and later two types of scale-free networks with varying degrees of clustering, with and without normalised payoffs (i.e. wealth inequality). Our findings suggest that the race tensions previously found in homogeneous networks are lowered, but that this effect only occurs in the presence of a certain degree of relational heterogeneity. In other words, spatial complexity by itself is not sufficient for the expectation of tempering the need for regulatory actions. Amongst all the network types studied, we {found} that scale-free networks with high clustering are the least demanding in terms of regulatory need, closely followed by regular scale-free networks. 

The questions of how network structures and diversity influence the outcomes of behavioural  dynamics, or the roles of  network reciprocity,  have been studied extensively in many fields, including Computer Science, Physics, Evolutionary Biology and Economics \citep{ahuja2000collaboration,Szabo2007,santos2006pnas,santos:2008:nature,ranjbar2014evolution,perc2013evolutionary,Hanijcai2018,perc2017statistical,raghunandan2012sustaining,perc2019social}.  Network reciprocity can promote the evolution of positive behaviours in various settings including cooperation dilemmas \citep{santos2006pnas,santos:2008:nature,ranjbar2014evolution,perc2017statistical}, fairness \citep{page2000spatial,szolnoki2012defense,wu2013adaptive} and trust \citep{Kumar_2020}. 
Their  applications are diverse, ranging  from healthcare \citep{newman2004coauthorship}, to network  interference and influence maximization  \citep{1500416,bloembergen2014influencing,cimpeanu2019exogenous},  and to climate change \citep{SantosPNAS2011}.
The present work  contributes new insights to this literature by studying the role of network reciprocity in the context of a technology development race.  
This strategy scenario is more intricate than the above-mentioned game theoretical scenarios (i.e., cooperation, trust and fairness) because, on the one hand, whether a social dilemma arises (where a collectively desired behaviour is not selected by evolutionary dynamics) depends on external factors  (e.g., risk probability $p_r$ in the early regime and monitoring probability $p_{fo}$ in the late regime)  \citep{han2019modelling}. On the other hand, the collectively desired behaviour in the arisen social dilemma is  different depending on the time-scale in which the race occurs. Interestingly, regardless of this more complex nature of the scenario,  the different desirable behaviours can always be promoted in heterogeneous networks.

As an avenue of exploring the role of prominent players in the development race, we make use of a previously proposed model of studying the influence of nodes based on their degrees of connectivity \citep{santos:2008:nature}. These highly connected individuals have a tendency towards safety compliance in comparison to their counterparts. In an attempt to exploit this effect, as well as to better understand the impact of such seemingly significant nodes, we introduced several pathological players \citep{pacheco2011messianic,Cardillo_2020, Kumar_2020} in key locations of the network (highly connected nodes). We showed the role of hubs in slowing development and promoting safety, and argue that a small minority of influential developers can drastically reduce race tensions in almost all cases. The addition of pathological participants in these important locations can play a key role in the emergence of safety, without sacrificing innovation, and this effect is robust under uncertain race conditions. {Our contribution explains the effects of heterogeneity in the networks that underlie the interactions between developers and teams of developers. We contend that there exist several ways in which this type of network heterogeneity could be promoted by relevant decision-makers, but argue that such mechanisms merit a dedicated body of research. Some examples of this could include dynamical linking \citep{pacheco2006dynamic}, whereby the relationship between two nodes could be altered by an outside decision-maker or the parties involved, or modifying the stakeholders' access to information, thereby amplifying selection dynamics \citep{Tkadlec2021}}.

{We note that our analyses focus on the binary extremes of developer behaviour, safe or unsafe development, in an effort to focus an already expansive problem into a manageable discussion. The addition of conditional, mixed, or random strategies could provide the basis for a novel piece of work. As observed with conditionally safe players in the well-mixed scenario \citep{han2019modelling}, we envisage that these additions would show little to no effect in the early regime, with the opposite being true for the late regime, at least in homogeneous settings.}

In short, our results have shown that heterogeneous networks can significantly mediate the tensions observed in a well-mixed world, in both early and late development  regimes \citep{han2019modelling}, thereby  reducing the need for regulatory actions. Since a real-world network of contacts among technological firms and developers/researchers appears to be highly non-homogeneous, our findings provide important insights for the design of technological regulation and governance frameworks (such as the one proposed in the EU White Paper \citep{EUAIwhitepaper2020}). Namely,  the underlying structure of the relevant network (among developers and teams) needs to be carefully investigated to avoid for example unnecessary actions (i.e. regulating  when that is not needed, as would have been otherwise suggested in homogeneous  world models).  Moreover, our findings suggest to increase heterogeneity or diversity  in the network as a way to escape tensions arisen from a race for technological supremacy. 

\section*{{Methods}}
{Below we describe different network structures and the details of how simulations on those networks are carried out. }
\subsection*{Population Dynamics}
 We consider a population of agents distributed on a network (see below for different network types), who are randomly assigned a strategy AS or AU.  
 At each time step or generation, each agent plays the game with its immediate neighbours. The {sucesss of each agent (i.e., its fitness) is the sum of the payoffs in all these encounters.} In the SI, we also discuss the limit where scores are normalised by the number of interactions (i.e., the connection {\it degree} of a node) \citep{santos2006new}. {Each individual fitness, as detailed below, defines the time-evolution of strategies, as successful choices will tend to be imitated by their peers.}
 
  At the end of each generation, a randomly selected  agent $A$ with {a fitness $f_A$  chooses to copy the strategy of a randomly selected neighbour, agent $B$, with fitness $f_B$ with a probability $p$ that increases with their fitness difference. Here we adopt the well-studied Fermi update or pairwise comparison rule, where } \citep{traulsen2006,santos2012role}: 
  {
      \begin{equation}
    	\label{fermi-function}
        p = (1+e^{\beta(f_A- f_B)})^{-1}.
     \end{equation}
     }
 In this case, $\beta$ conveniently describes the selection intensity --- i.e., the importance of individual success in the imitations process: $\beta=0$ represents neutral drift while $\beta \rightarrow \infty$ represents increasingly deterministic imitation \citep{traulsen2006}. Varying $\beta$ allows capturing a wide range of update rules and levels of stochasticity, including those used by humans, as measured in lab experiments \citep{zisisSciRep2015,randUltimatum,grujic2020people}. In line with previous works and lab experiments, we set $\beta = 1$ in our simulations, ensuring a high intensity of selection \citep{pinheiro2012selection}. {This update rule implicitly assumes an asynchronous update rule, where at most one imitation occurs at each time-step. We have nonetheless confirmed that similar results are obtained with a synchronous update rule.}

\subsection*{Network Topologies}

{{Links} in the network describe a relationship of proximity both in the interactional sense (whom the agents can interact with), but also observationally (whom the agents can imitate). Ergo, the network of interactions coincides with the imitation network \citep{ohtsuki2007breaking}. As each network type converges at different rates and naturally presents with various degrees of heterogeneity, we choose different population sizes and maximum numbers of runs in the various experiments to account for this while optimising run-time.}
  
  Specifically, to study the effect of network structures on the safety outcome, we will analyse the following types of  networks, from simple to more complex: 
 \begin{enumerate} 
   \item Well-mixed population (WM) (complete graph): each agent interacts with all other agents in a population,
  \item  Square lattice (SL) of size $Z = L \times L$ with periodic boundary conditions--- a  widely adopted population structure in population dynamics and evolutionary games (for a survey, see \citep{Szabo2007}). Each agent can only interact with its four immediate edge neighbours. We also study the 8-neighbour lattice for confirmation (see SI),  
  \item Scale-free (SF) networks \citep{barabasi1999emergence,dorogovtsev2010complex,newman2003structure}, generated through two growing network models --- the widely-adopted Barab\'{a}si-Albert (BA) model \citep{barabasi1999emergence,albert2002} and the Dorogovtsev-Mendes-Samukhin (DMS) model \citep{dorogovtsev2001size,dorogovtsev2010complex}{, the latter of which} allows us to assess the role of a large number of triangular motifs (i.e. high clustering coefficient). Both BA and DMS models portray a power-law degree distribution $P(k)\propto k^{-\gamma}$ with the same exponent $\gamma=3$. In the BA model, graphs are generated via the combined mechanisms of growth and preferential attachment where new nodes preferentially attach to $m$ existing nodes with a probability that is proportional to their already existing number of connections \citep{barabasi1999emergence}. In the case of the DMS model, new connections are chosen based on an edge lottery: each new vertex attaches to both
ends of randomly chosen edges, also connecting to $m$ existing nodes. As such, we favour the the creation of triangular motifs, thereby enhancing the clustering coefficient of the graph. In both cases, the average connectivity is $z = 2m$.  
 \end{enumerate}  

Overall, WM populations offer a convenient baseline scenario, where interaction structure is absent. With the SL we introduce a network structure, yet one where all nodes can be seen as equivalent. Finally, the two SF models allow us to address the role of heterogeneous structures with low (BA) and high (DMS) clustering coefficients. The SF networks portray a heterogeneity which mimics the power-law distribution of wealth (and opportunities) of real-world settings. 

\subsection*{Computer Simulations}

For well-mixed populations and lattice networks, we chose populations of $Z = 100$ agents and $Z = 32 \times 32$ agents, respectively. In contrast, for scale-free networks, we chose $Z = 1000$, while also pre-seeding with agents 10 different networks (of each type) on which to run all the experiments in an effort to minimise the effect of network topology and the initial, stochastic distributions of players. We chose an average connectivity of $z = 4$ for our SF networks, to coincide with the regular average connectivity in square lattices for the sake of comparison. 

We simulated the evolutionary process for $10^4$ generations (a generation corresponds to $Z$ time-steps) in the case of scale-free networks and $10^3$ generations otherwise. 
The equilibrium frequencies of each strategy were obtained by averaging over the final $10^3$ steps. Each data point shown below was obtained from averaging over 25 independent realisations, for each of the 10 different instances used in each network topology.

\section*{Data Availability}
Code for simulations is available upon request. 

\section*{Acknowledgements}
T.A.H., L.M.P., T.L. and T.C., are supported by Future of Life Institute grant RFP2-154. T.C. and T.A.H. also acknowledge support from Berkeley Existential Risk Initiative (BERI) collaboration fund. LM.P. support by NOVA-LINCS (UIDB/04516/2020) with the financial support of FCT-Funda\c c\~ao para a Ci\^encia e a Tecnologia, Portugal, through national funds. 
F.C.S. acknowledges support from FCT Portugal (grants UIDB/50021/2020, PTDC/MAT-APL/6804/2020 and  PTDC/CCI-INF/7366/2020). T.L benefits from the support by the Flemish Government through the AI Research Program and F.C.S and T.L.by TAILOR, a project funded by EU Horizon 2020 research and innovation program under GA No 952215. T.L. is furthermore supported by the F.N.R.S. project with grant number 31257234,  the F.W.O. project with grant nr. G.0391.13N, the FuturICT 2.0
(\url{www.futurict2.eu}) project funded by the FLAG-ERA JCT 2016 and the Service Public de Wallonie Recherche under grant n\textdegree\ 2010235–ARIAC by DigitalWallonia4.ai.  T.A.H. is also supported by a Leverhulme Research Fellowship (RF-2020-603/9).

\section*{Author contributions}
T.C., F.C.S., L.M.P., T.L. and  T.A.H.  designed the research; T.C. implemented the software and prepared the figures; T.C., F.C.S., L.M.P., T.L. and  T.A.H. wrote the manuscript; T.C., F.C.S., L.M.P., T.L. and  T.A.H. reviewed the manuscript.

\section*{Competing interests}
The authors declare no competing interests.



\renewcommand{\thefigure}{S\arabic{figure}}
 \renewcommand{\thetable}{S\arabic{table}}
 \setcounter{figure}{0}   

\newpage

\section*{Additional simulation results}
\label{section:supplementary}

\begin{figure}[H]
	\centering
	\includegraphics[width=0.8\linewidth]{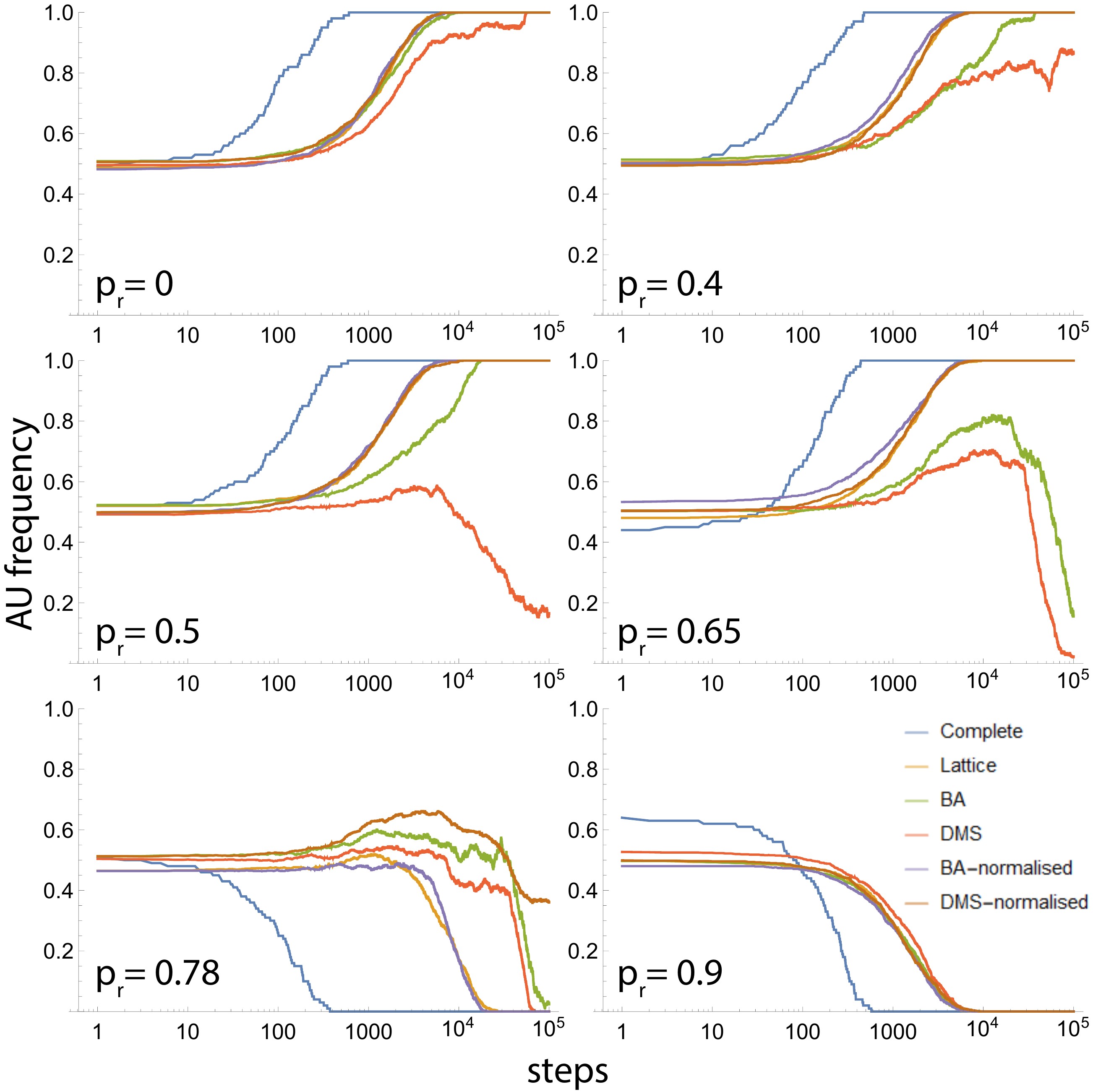}
	\caption{Scale-free networks (especially highly clustered networks) reduce unsafe behaviour in the dilemma regions of the early race, shown using typical runs for different risk probability values, for each type of network. Parameters: $c = 1, \ b = 4, \ B = 10^4, \ \beta = 1.$}
	\vspace{-0.5cm}
	\label{fig:evol-plots}
\end{figure}

To further illustrate the key differences between each type of network, we plot typical simulation runs for different $p_r$ risk probability values in the area (\textbf{II}) of the early AI race (see Figure \ref{fig:evol-plots}). It is immediately apparent that the two un-normalised scale-free networks provide significant improvements in safety compliance in the dilemma zone. This is further compounded by the effect of clustering on the threshold at which safe development becomes evolutionarily stable. Specifically, we note that when the risk of a disaster occurring due to inadequate safety compliance is intermediate (see, e.g. $p_r = 0.5$ and $0.65$), we see a definitive improvement in highly clustered networks (i.e. DMS) as opposed to the basic BA model.

\begin{figure}
	\centering
	\includegraphics[width=0.4\linewidth]{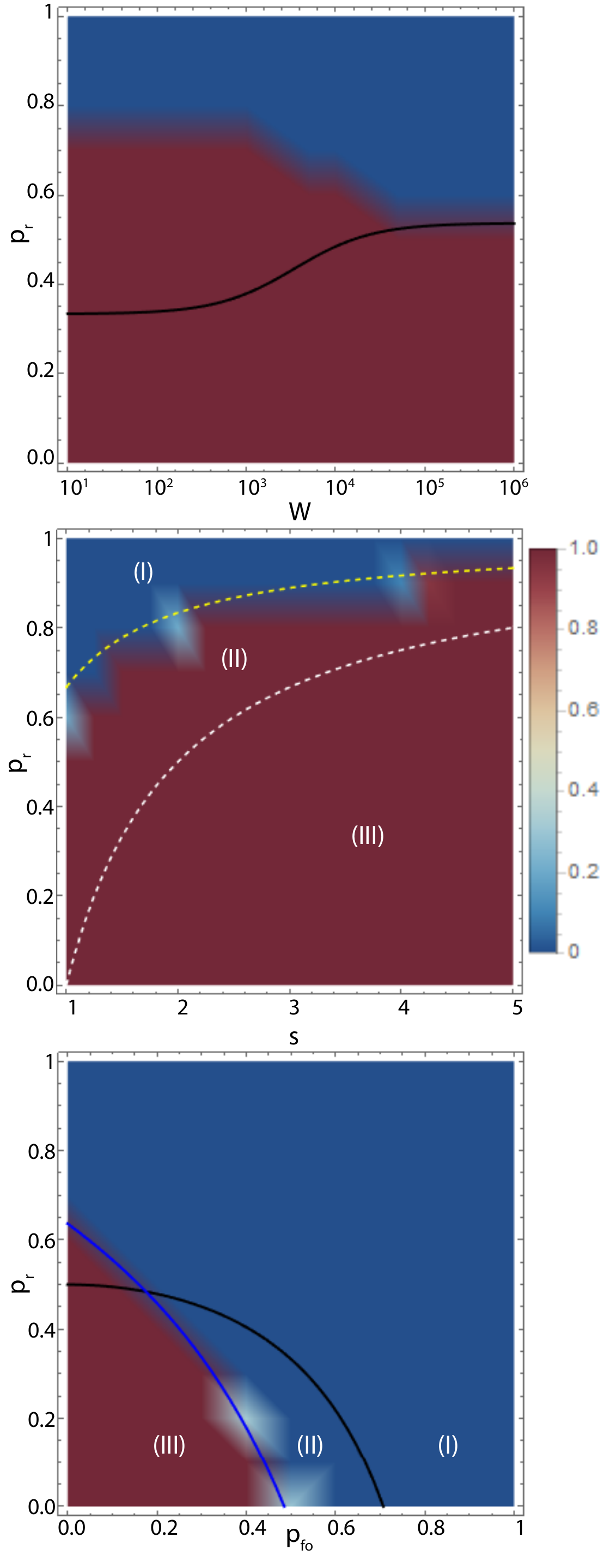}
	\caption{Total AU frequencies for the 8-neighbours lattice. The top row reports the spectrum between an early and late AI race (for varying $W$, with $p_{fo} = 0.1, \ s = 1.5$),  the middle row addresses the early regime for varying $s$ and $p_r$ ($p_{fo} = 0.5, \ W = 100$), and the bottom row addressees the late regime for varying $p_{fo}$ and $p_r$ ($s = 1.5$, $W = 10^6$). Other parameters: $ \ c = 1, \ b = 4, \ B = 10^4, \ \beta = 1.$}
	\vspace{-0.5cm}
	\label{fig:lattice-eight}
\end{figure}

Figure \ref{fig:lattice-eight} confirms the similar trends encountered in the regular square lattice. There are some very minor differences, but there is very little difference between well-mixed, the normal four-neighbour lattice and the eight-neighbour lattice. We confirm the similar late convergence found previously in some cases of the regular lattice. 

We see very few improvements over the previously mentioned results on homogeneous populations. Interestingly, there is an area in the late regime where this type of normalised scale-free network produces more unsafe results (undesirably so) than either the well-mixed or lattice variants. We see some slight improvements in area \textbf{(II)} of the early regime. 

\begin{figure}
	\centering
	\includegraphics[width=\linewidth]{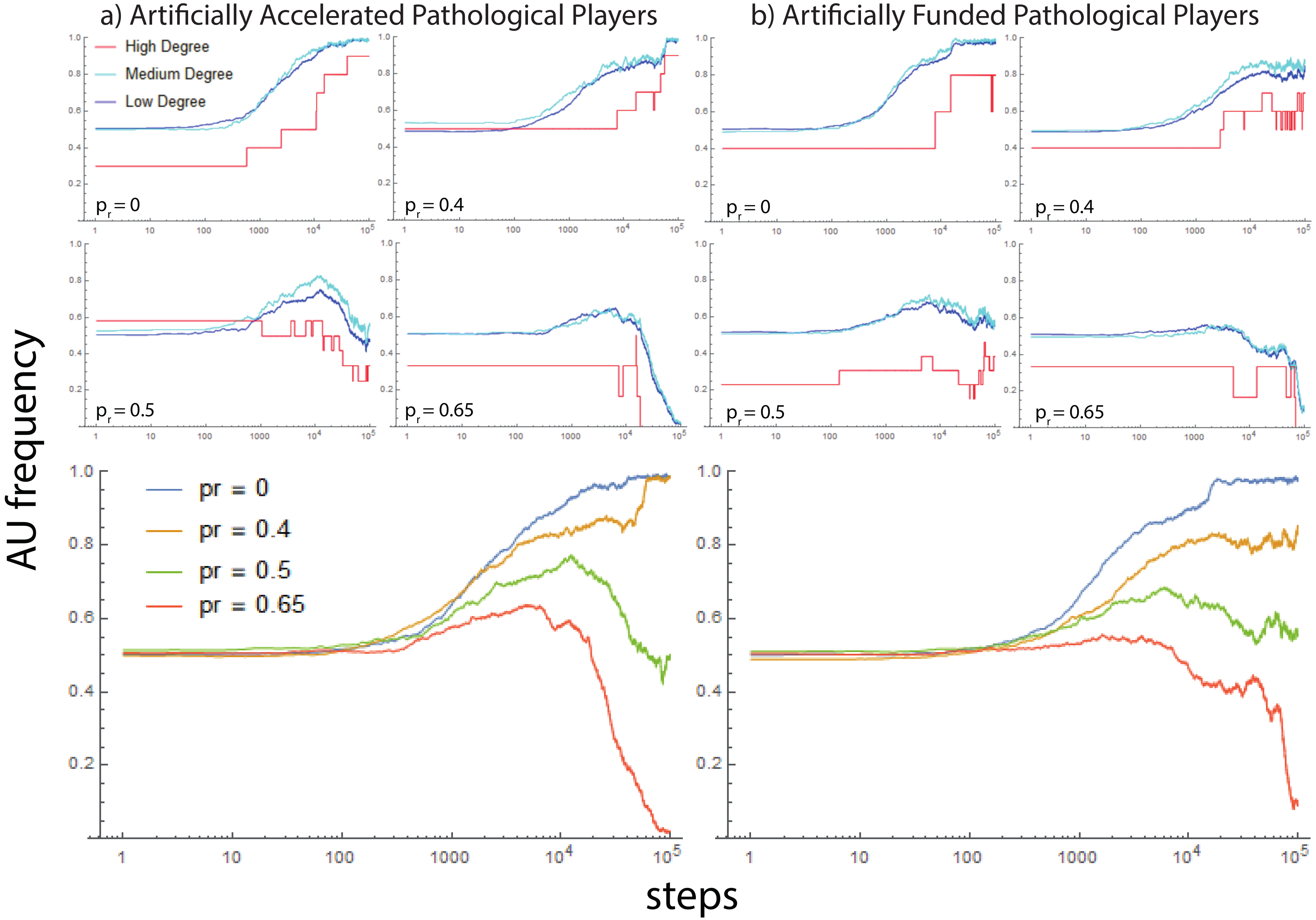}
	\caption{Typical runs exploring the evolutionary degree distribution of unsafe behaviour in an early AI race, following the artificial acceleration (or funding) of safety zealots (pathological safe players) in the population interacting in DMS networks. We randomly allocate 10\% of high degree individuals as safety zealots. Note that we measure the frequency for the whole population, including the pathological players. Parameters: $c = 1, \ b = 4, \ s = 1.5, \ B = 10^4, \ W = 100, \ \beta = 1.$}
	\label{fig:interference-pathological}
\end{figure}

In order to better understand the role and influence of highly connected zealots in the population, as well as to explore any potential for a government or regulatory agency to interfere in the AI race, we artificially accelerate or fund the safety zealots that had been introduced previously. For this analysis, we choose a small number ($10\%$ of high-degree nodes) of individuals, to check whether a very small minority can be exploited by an external investor. In addition to the introduction of  players following pathological safe behaviour, we either accelerate their development (similarly to how unsafe players gain increased speed, in this case we add $\frac{sB}{W}$ to the influential pathological players' payoffs, where $s = 2$), or heavily invest in these players (to the extent that other players will always imitate them, by increasing their payoffs by a very large amount $10^7$). Figure \ref{fig:interference-pathological} displays our findings - with very slight improvement throughout. Each approach has its merits in different regions of the early regime, and we see the effectiveness of funding highly connected nodes when the risk for disaster is low. On the other hand, a high risk improves the efficacy of speeding up the development for these dedicated minorities. We note that targeting a very small minority of highly influential players is not sufficient to mitigate the race tensions entirely. Further exploration on this topic would provide more insight into how external interference can be deployed efficiently.

\begin{figure}
	\centering
	\includegraphics[width=0.80\linewidth]{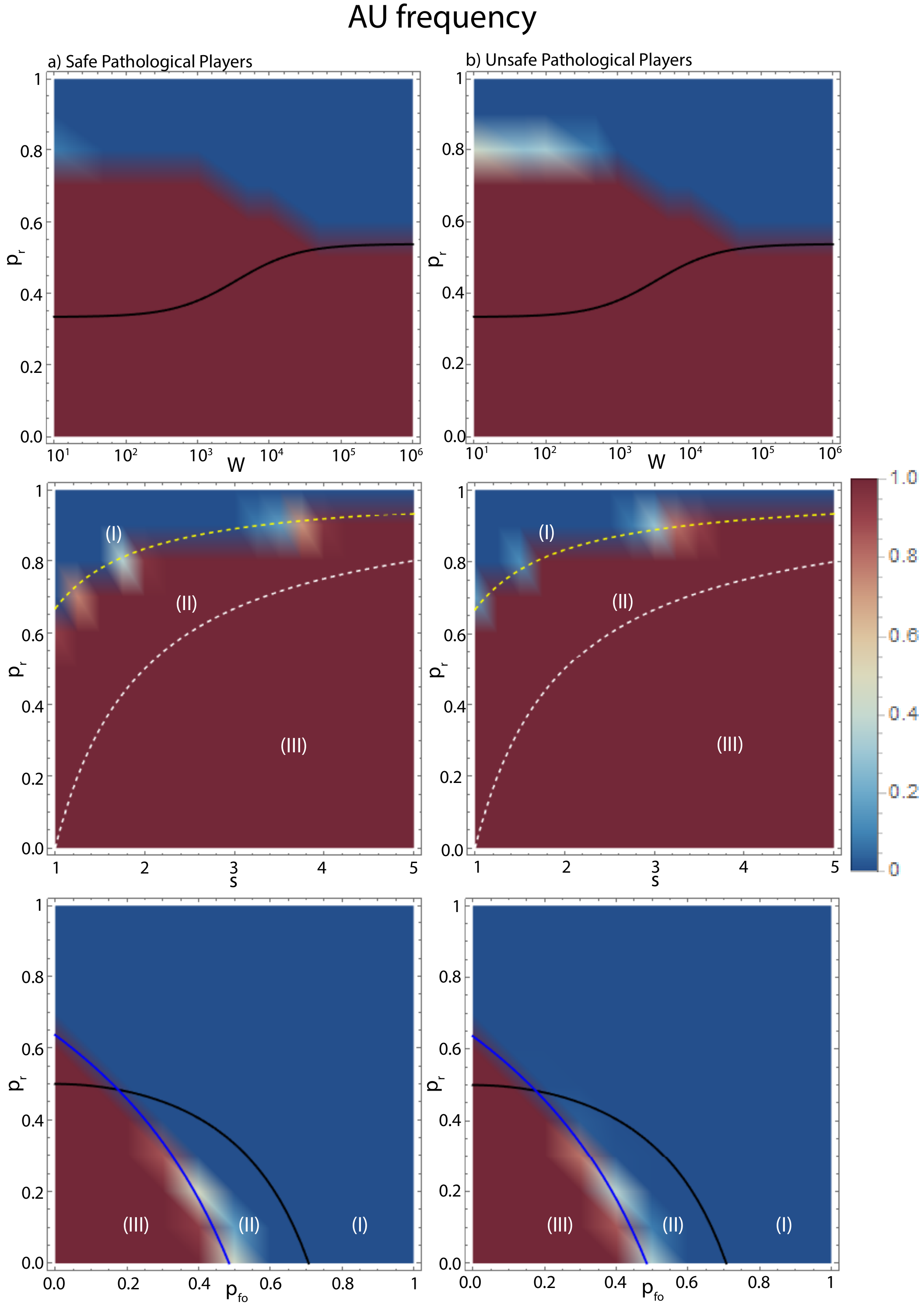}
	\caption{Introducing safe and unsafe zealots in the well-mixed scenario. Please note that the pathological players are excluded from these frequencies. The top row reports the spectrum between an early and a late AI race ($p_{fo} = 0.1, \ s = 1.5$), the middle row addresses the early regime in more detail ($p_{fo} = 0.5, \ W = 100$) and the bottom row considers a late AI race ($W = 10^6, \ s = 1.5$). Parameters: $ c = 1, \ b = 4, \ B = 10^4, \ \beta = 1.$}
	\label{fig:well-mixed-pathological}
\end{figure}

We study a comprehensive view of pathological players (zealots) planted in a well-mixed network (see Figure \ref{fig:well-mixed-pathological}), but in this case modifying $10\%$ of the total population (not just highly connected nodes). We remove the pathological players from the frequency average to show how these affect the remainder of the population. We see very little effect of pathological players and we suggest that much lower $\beta$ values would be required to see an effect. With the addition of mutation and more stochasticity, it would be possible for these pathological players to have a significant impact on the outcome. 

\begin{figure}
	\centering
	\includegraphics[width=0.95\linewidth]{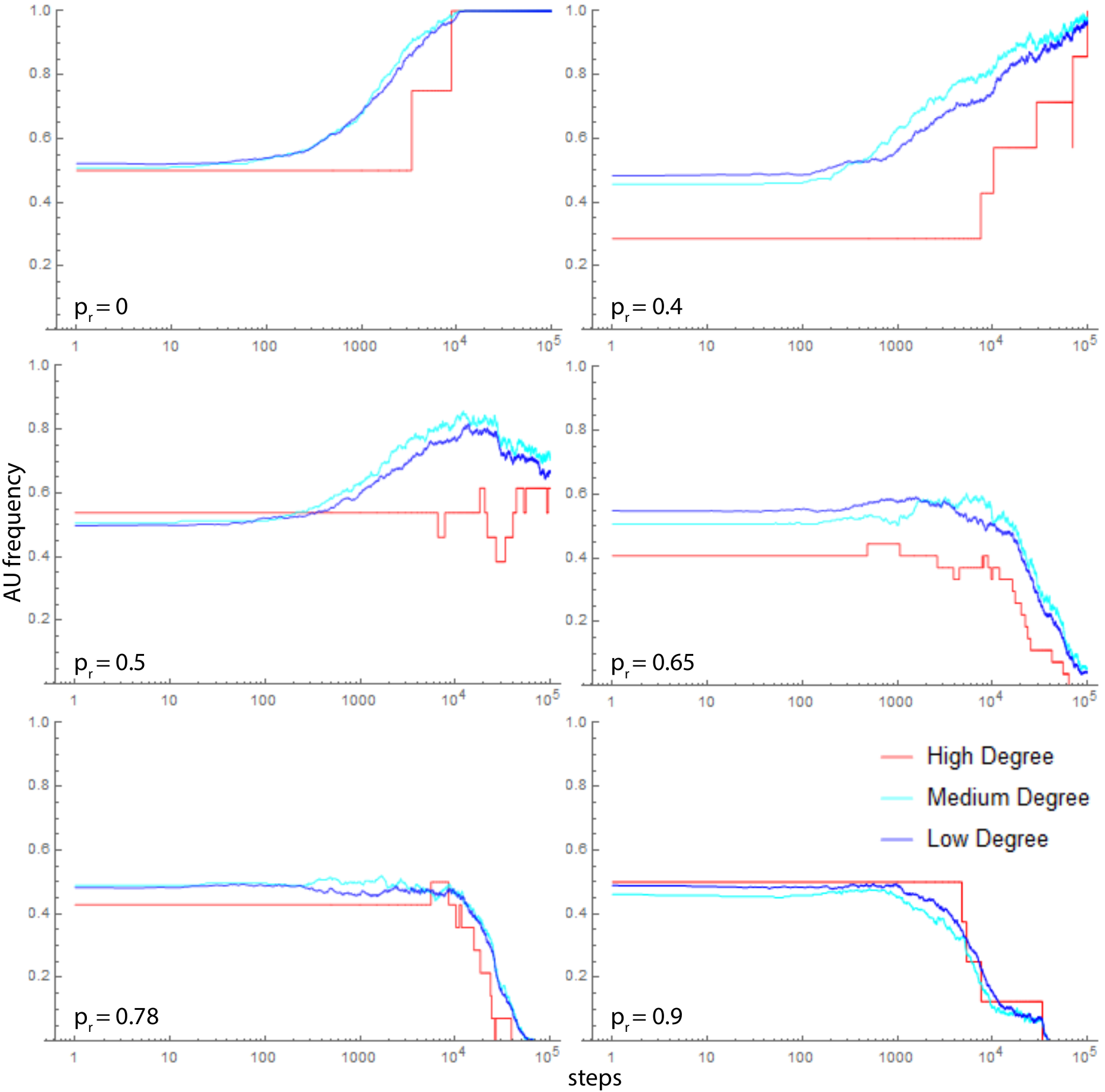}
	\caption{Typical runs showing the distribution of unsafe behaviour (AU) in an early AI race, grouped by degree class (connectivity) of the nodes on DMS networks, for different risk probabilities. Parameters: $c = 1, \ b = 4, \ s = 1.5, \ B = 10^4, \ W = 100, \ \beta = 1.$}
	\label{fig:degree-evolution}
\end{figure}

Figure \ref{fig:degree-evolution} shows the evolution over time of unsafe behaviour (AU)  in the dilemma zone of an early AI race for different environments (corresponding to varying probability values of a disaster caused by insufficient safety regulation, $p_r$). High-degree individuals appear to have a higher tendency towards safety compliance (at equilibrium) when compared to their lowly or moderately connected counterparts, except for region (\textbf{III}), where highly connected individuals are driving to innovate (optimally so). {In spite of this}, we see the same trends for regions (\textbf{I}) and (\textbf{III}). However, in  region (\textbf{II}), highly connected individuals become important leaders in the shift from unsafe to safe behaviour in the AI race. Specifically, for large $p_r$ values (see $p_r = 0.65; \ p_r = 0.78$), there is an evident disparity between the high degree individuals and the bulk of the population, and indeed, this is the region in which heterogeneity improves safety compliance the most. For low $p_r$ values, heterogeneity fails to improve the outcome, but it does serve as an equaliser for intermediate risk values ($p_r = 0.5$). Regulatory actions would therefore still be required to {constrain developers} when heterogeneity cannot improve safety enough in region \textbf{II}, in the case of low risk of disaster to occur. 

\begin{figure}
	\centering
	\includegraphics[width=0.77\linewidth]{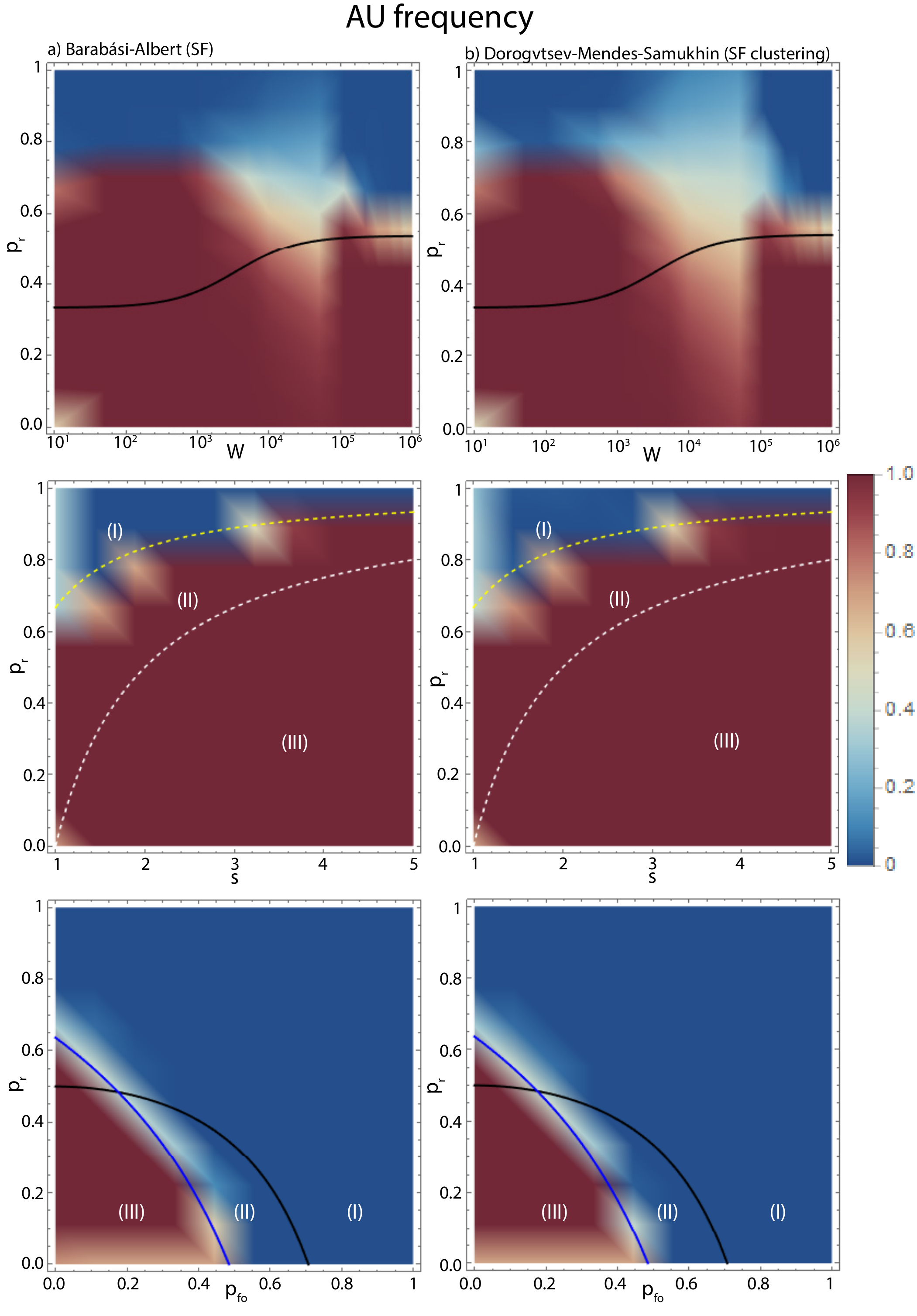}\vspace{-0.25cm}
	\caption{Comparison between the two different scale-free networks, BA and DMS. In this case, the payoffs have been normalised.  The top row reports the spectrum between an early and a late AI race ($p_{fo} = 0.1, \ s = 1.5$), the middle row addresses the early regime in more detail ($p_{fo} = 0.5, \ W = 100$) and the bottom row considers a late AI race ($W = 10^6, \ s = 1.5$). Parameters: $ c = 1, \ b = 4, \ B = 10^4, \ \beta = 1.$}
	\label{fig:scale-free-normalised}
	\vspace{-0.5cm}
\end{figure}


\begin{figure}
	\centering
	\includegraphics[width=0.9\linewidth]{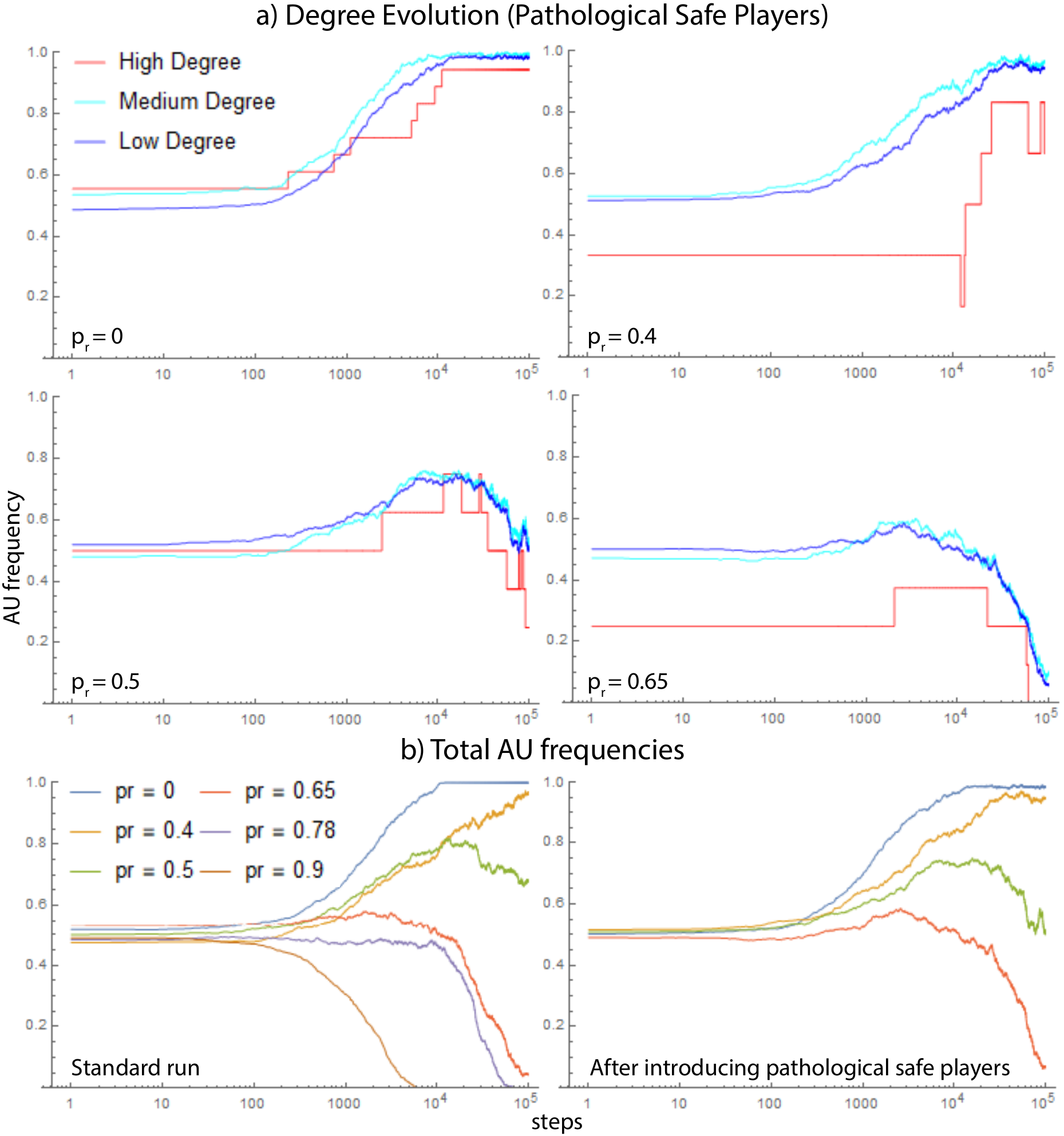}
	\caption{Typical runs exploring the evolutionary degree distribution of unsafe behaviour in an early AI race, following the introduction of safety zealots (pathological safe players) in the population of DMS networks. We randomly allocate 10\% of high degree individuals as safety zealots. Note that we measure the frequency for the whole population, including the pathological players. Parameters: $c = 1, \ b = 4, \ s = 1.5, \ B = 10^4, \ W = 100, \ \beta = 1.$}
	\label{fig:pathological-safe-degree-evolution}
\end{figure}

\begin{figure}
	\centering
	\includegraphics[width=\linewidth]{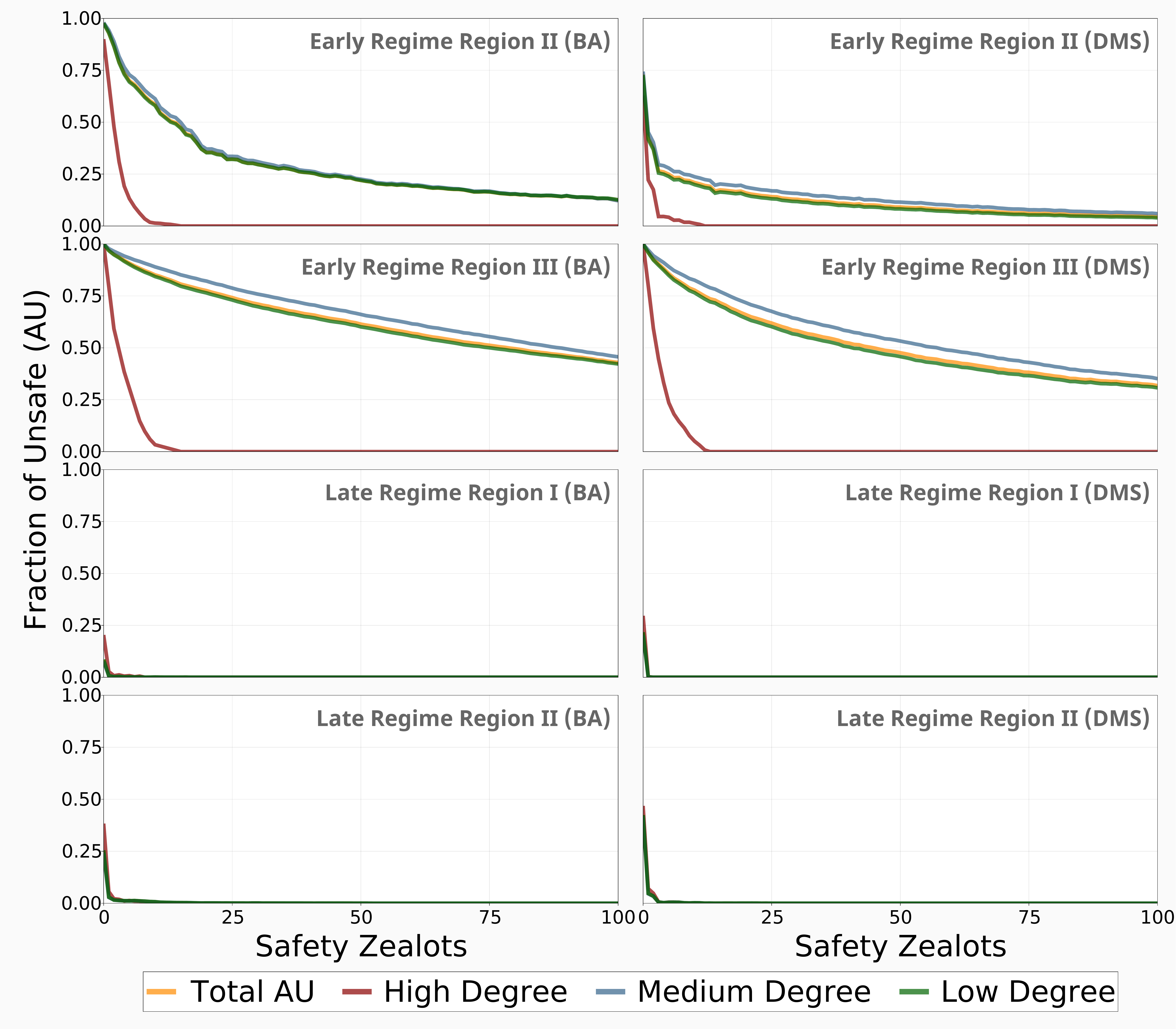}
	\caption{Hubs prefer slower, thus safer developments in the early race, and this can be further exploited by progressively introducing safety zealots in highly connected nodes. We show the results for both regimes, as well as the appropriate regions where safety (early region II and late region I), and conversely where innovation (early region III and late region II) are the preferred collective outcomes. The top four panels report the results for the early regime ($p_{fo} = 0.5, \ W = 100$ with $p_r = 0.5$ for region II and $p_r = 0.1$ for region III), and the bottom four do so for the late regime ($p_{fo} = 0.6, \ W = 10^6$ with $p_r = 0.3$ for region I and $p_r = 0.1$ and region II).  Other parameters: $c = 1, \ b = 4, \ B = 10^4, \ s = 1.5, \  \beta = 1.$}
	\label{fig:zealots-full}
\end{figure}

\begin{figure}
	\centering
	\includegraphics[width=\linewidth]{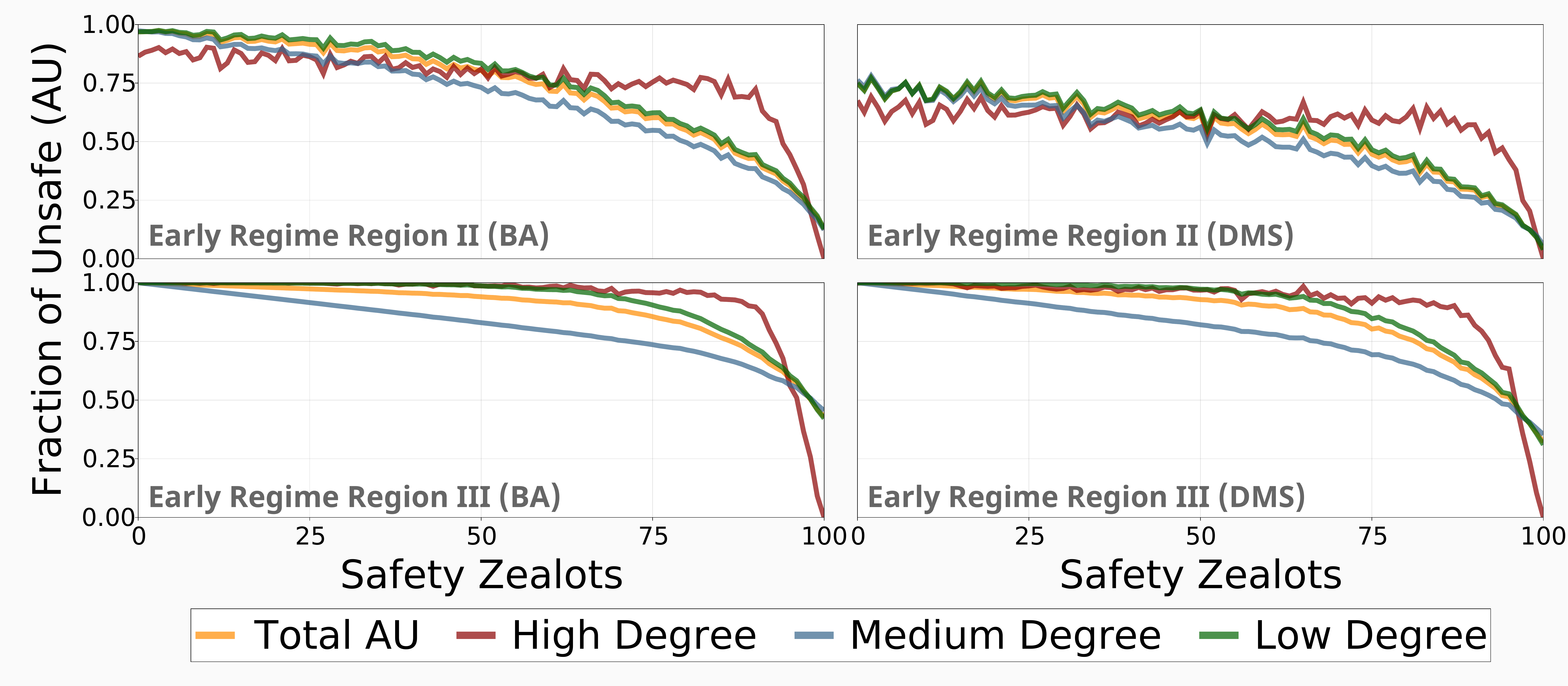}
	\caption{Introducing safety zealots in reverse order (still selecting the top 10\% of nodes based on degree connectivity) does not produce the same exponential increase in safety that we had seen in Figure 3. We show the results for the early regime, as well as the appropriate regions where safety (region II), and conversely where innovation (region III) are the preferred collective outcomes. Parameters are $p_r = 0.5$ for region II and $p_r = 0.1$ for region III, chosen for clear presentation.  Other parameters: $c = 1, \ b = 4, \ B = 10^4, \ s = 1.5, \  \beta = 1, \ p_{fo} = 0.5, \ W = 100.$}
	\label{fig:zealots-reversed}
\end{figure}

\end{document}